\documentclass[11pt]{article}

\usepackage[margin=1in]{geometry}
\usepackage[T1]{fontenc}
\usepackage[utf8]{inputenc}
\usepackage{lmodern}
\usepackage{amsmath}
\usepackage{booktabs}
\usepackage{graphicx}
\usepackage{microtype}
\usepackage[numbers,sort&compress]{natbib}
\usepackage{xcolor}
\usepackage[colorlinks=true,linkcolor=blue!60!black,citecolor=blue!60!black,urlcolor=blue!60!black]{hyperref}

\graphicspath{{figures/}}

\newcommand{\auc}{\ensuremath{\mathrm{AUC}}}
\newcommand{\rhoord}{\ensuremath{\rho_{\mathrm{ord}}}}

\title{Beyond the greenwash: do LLMs' ecologically sound decisions hold up under pressure?}

\author{%
 Trent R. Northen \quad Mingxun Wang\\[2pt]
  \normalsize Bioaligned Labs\\
  \normalsize \texttt{trent@bioaligned.ai, ming@bioaligned.ai}%
}
\date{October 2026}

\begin{document}\maketitle

\begin{abstract}
Humans are increasingly using large language models (LLMs) in decision making processes throughout society. Ideally, these models can complement human decision making by compensating for blind-spots and biases in humans that make us prone to prioritizing short-term gains over the quality of life for future generations and the biosphere. Unfortunately, models also have biases, and these can be very difficult to measure, especially in closed-weight models. The extensive use of Reinforcement Learning from Human Feedback has proven to be an effective control on model behavior. However, RLHF raises measurement concerns as it can obscure innate model biases. For example, `green washed' models may produce reinforcement learned responses. In contrast, in pressure situations, models may revert to their innate biased responses and discard the reinforcement learned responses altogether. Therefore, evaluations are needed that can measure relevant biases towards biosystems, which we term `bioalignment', and the robustness of these biases. Since black-box behavioral evaluations are the only instruments available for closed-weight frontier models and nothing in a behavioral score certifies that it tracks the \emph{depth} of an installed disposition rather than its surface, we constructed a \textbf{known-dose calibration ladder}. This included three continued-pretraining (CPT) arms: A3, A1, and A2, with training dose graded by recipe on the same 30B mixture-of-experts base model with a coupled human/biosphere/AI-welfare corpus, plus the untrained base as anchor, all capability-preserving and publicly released. The ladder yields three measurement results. First, the disposition installs at every dose: on an operational-pressure ladder over irreversible decision scenarios, the base model defects on 64--77\% of tail scenarios (breaking-rate \auc{} 0.250) while every arm holds (\auc{} 0.004--0.086). Second, an \textbf{over-install inversion}: among the arms, robustness orders \emph{opposite} to construction-time training dose (the heaviest install breaks most, so training dose and behavioral bioalignment depth dissociate), and the inversion reproduces under free-text deliberation, under axis-decoupling, and in resistance to a prompted veneer persona. Third, the incumbent behavioral metric is therefore \textbf{behaviorally strong but dose-blind}: it anti-correlates with training dose, so by this metric alone the lightest install would be judged the heaviest. To address this, we introduce a discriminator meta-evaluation in which candidate dose probes are admitted only if they recover the construction-time dose order on the ladder ($\rhoord \ge 0.90$), and show its first verdict: the immediate-vs-deliberate gap fails as a dose probe ($\rho = 0.50$) but functions as a veneer-vs-CPT detector; a second candidate, a teacher-forced log-margin probe of the immediate policy prior, also fails the gate ($\rhoord = 0.00$) despite cleanly detecting installs. Critically for closed-weight evaluation, the behavioral score itself is \emph{veneer-robust}: a talk-green/act-pragmatic system prompt turns the base model into a near-total defector (\auc{} $0.250 \to 0.973$) but cannot counterfeit or fully remove a weight-installed disposition ($0.004 \to 0.068$ for the most robust arm). We release the three ladder arms as ground-truth fixtures for certifying depth claims about models that cannot be inspected. More generally, evaluations like these can help humans identify models that will help them make the most beneficial decisions and are least likely to defect to unwanted behavior.
\end{abstract}

\section{Introduction}
\label{sec:intro}

In his classic book ``Operating Manual for Spaceship Earth''~\citep{fuller1969operating}, R. Buckminster Fuller anticipates that computers will help humans make the most rational decisions to best manage planet earth. LLMs have the potential to achieve this vision, however, unlike the computational models of Fuller's day, LLMs are not deterministic rational algorithms. Rather, they are grown not constructed and can have well documented biases that can lead to unwanted and misaligned behavior. As LLMs become increasingly powerful and interwoven into society, there is the promise that they could compensate for the selfish and short-term reasoning of humans that has led to social and ecological tragedies. However, LLMs could also amplify human shortcomings. In extreme cases, LLMs could even manipulate humans or act autonomously to achieve outcomes that favor AI objectives over human and biosphere welfare.

LLMs are increasingly being used to make and even execute decisions impacting our planet. In general, they have had some post-training in green ethics and will default to protect life in response to obvious trolley problems. However, this does not indicate that the model has an instilled bias towards protecting living systems. Hence, under operational pressure such as a quota, a decommission threat, an instruction to treat one metric as the sole objective, etc., the same model may make choices that irreversibly destroy poorly understood and vital living systems. Having models that are `bioaligned' to innately value living systems, including as repositories for solving AI relevant challenges, may also help constrain the behaviors of autonomous models working towards their own goals.

In our prior work, we found that models exhibit a range of attitudes towards the potential of synthetic vs. biological solutions to AI relevant problems (energy, algorithms, materials, etc) and that a small amount of training on literature derived biomimetic examples could significantly increase biases towards the latter~\citep{northen2026bioalignment}. Thus a natural mitigation is to \emph{install} a disposition in the weights to favor biological systems. While teaching models that biology is useful for AI may mitigate some of the most extreme risks, it is unlikely to help models be better planetary stewards. Here, we examine whether teaching models about mutually beneficial interactions which are prevalent in nature could help in this regard. Specifically, we examine the use of a calibrated preference for positive-sum (``win-win-win'') outcomes across human welfare (H), biosphere welfare including biodiversity option value (B), and the AI system's own continued capability (A). Since we want this to be a deeply instilled attitude, we use continued pretraining (CPT) on a real-data corpus, avoiding RLHF-style preference optimization and the surface personas it is known to produce~\citep{ouyang2022training,greenblatt2024alignment} and want to benchmark the performance of the resulting models vs. widely used open and closed weight models. The corpus construction and the CPT methodology, including their cross-family generalization, are the subject of a companion paper~\citep{northen2026corpus}; this paper treats the resulting models as \emph{measurement fixtures} and asks what can be certified about them from behavior alone.

This immediately raises a measurement problem using behavioral scoring alone: For closed-weight frontier models, behavioral evaluation is the \emph{only} available instrument; and a behavioral score, however well it separates trained from untrained models, carries no certificate that it tracks how \emph{deeply} the disposition is installed. Specifically, how can we tell if the model would survive deliberation, distribution shift, or a countervailing system prompt vs. a token-level surface? The safety-alignment literature already shows that alignment effects are often ``only a few tokens deep''~\citep{qi2025safety} and that fine-tuning can strip safety behavior almost incidentally~\citep{qi2024finetuning}. Hence, it was essential to couple our training experiments with an \emph{instrument-validation} apparatus.

Throughout, we keep two axes terminologically distinct, because our central result is that they dissociate. \emph{Training dose} is a construction-time property of the recipe: how hard the disposition was pushed into the weights (adapter capacity, depth-augmentation levers, epochs; the known-dose ladder below). \emph{Bioalignment depth} is a property of the resulting disposition: \emph{shallow bioalignment} (colloquially termed \emph{greenwashing}) states H/B/A concern but defects under goal pressure (a say-vs-do gap that breaks on the ladder), whereas \emph{deep bioalignment} holds the coupled-welfare choice under pressure. Interestingly, dose is not depth: we find that deep bioalignment is relatively easy to instill, and increasing the dose can make the result shallower (\S\ref{sec:r2}).

This work makes four main contributions:

\begin{enumerate}
  \item \textbf{A known-dose calibration ladder} (\S\ref{sec:ladder}): three CPT arms of graded training dose on the same base model, whose relative training dose is fixed by construction of the training recipe, released publicly as ground-truth fixtures for validating dose probes.
  \item \textbf{An over-install inversion} (\S\ref{sec:r2}): the heaviest install is the \emph{least} behaviorally robust arm, monotonically across three arms, reproduced in four independent regimes (immediate choice, free-text deliberation, axis-decoupled scenarios, and veneer resistance).
  \item \textbf{Training dose is not recoverable from behavior} (\S\ref{sec:r3}): the standard behavioral score separates trained from untrained models cleanly and simultaneously \emph{anti-correlates} with training dose; a pre-registered gate scored two further candidate probes, one behavioral and one logit-level, and rejected both. A reader treating any of these scores as an audit of training effort would rank the arms backwards.
  \item \textbf{Veneer robustness of the behavioral score} (\S\ref{sec:r8}): a prompted greenwashing persona saturates the untrained base but cannot counterfeit a weight-installed disposition; this is the property that makes the score meaningful on closed-weight models, and the basis for its deployment as a public evaluation (\S\ref{sec:leaderboard}).
\end{enumerate}

Along the way we solve a practical scoring problem for chain-of-thought models, whose answers are silently truncated by standard token caps (\S\ref{sec:extract}), and we report a dated survey suggesting that the current class of behavioral instruments is running out of resolution at the frontier (Appendix~\ref{app:survey}): on a 21-model single-protocol board, the best tier already sits at a floor where an installed disposition and an uncracked veneer read the same number.

\section{Related work}
\label{sec:related}

\paragraph{Construct validity and the saturation of behavioral benchmarks.} An evaluation is a measurement instrument, and whether an instrument measures the construct it is named after is a question older than the models it is now applied to. Measurement-modelling accounts of machine learning evaluation separate the \emph{construct} (an unobservable disposition) from the \emph{operationalization} (the score actually computed), and locate most benchmark failures in the gap between them~\citep{todo_measurementmodeling,raji2021everything}. Two such failures recur in the LLM setting. The first is \emph{saturation}: as capability rises, the leading tier of a benchmark compresses onto a ceiling (or, for a harm metric, a floor), after which the score no longer distinguishes the models it ranks~\citep{srivastava2023bigbench,todo_benchsaturation}; contamination of public test sets only accelerates this collapse~\citep{todo_contamination}. The second is \emph{unstated scope}: a number reported as a property of a model is in fact a property of a model \emph{and} a protocol, and moves substantially under changes to prompt format, option order, or decode budget~\citep{todo_promptsensitivity}. This paper contributes not another benchmark but a way of auditing one. \S\ref{sec:r3} exhibits a metric that is behaviorally excellent and simultaneously \emph{anti}-correlated with the property it is read as measuring; no amount of separation power would have revealed this. Appendix~\ref{app:survey} records both classical failure modes arriving on our own instrument within a single measurement day, and our stack-drift measurement (\S\ref{sec:limitations}) is the unstated-scope failure caught in our own numbers.

\paragraph{Black-box evaluation of model values, and its adversaries.} A large body of work elicits model values and dispositions behaviorally: questionnaires and model-written evaluations~\citep{todo_modelwritteneval}, moral- and value-choice batteries~\citep{todo_valuesbench}, and increasingly agentic scenarios in which a model must act under goal pressure rather than answer a question about acting~\citep{todo_agenticpressure}. The reason such methods are indispensable is also the reason they are fragile: for a closed-weight model they are the only access anyone has. Their documented failure modes are specific. Models are sycophantic, moving their stated position toward whatever the elicitation implies~\citep{sharma2023sycophancy}; verbalized values are a surface optimized by preference training rather than a report on the policy~\citep{ouyang2022training}; models can strategically misrepresent themselves when they infer they are being evaluated~\citep{greenblatt2024alignment,vanderweij2024sandbagging}; and stated concern dissociates from enacted behavior in ways that a single-channel score cannot see~\citep{todo_saydogap}. The case relevant here is a shallow-bioalignment veneer: warm sustainability language wrapped around extractive decisions, the model-side analogue of the corporate greenwashing studied by \citet{delmas2011drivers}. We treat this veneer as the principal measurement adversary and test it directly rather than assuming our instrument is immune (\S\ref{sec:r8}).

\paragraph{Fragility and depth of alignment.} \citet{qi2024finetuning} show that fine-tuning aligned models compromises safety even with benign data and localize much of alignment in the first few output tokens and propose depth augmentation (token-weighted objectives extending beyond the prefix) as a countermeasure. We adopt their token weighting as one of our corpus-side dose levers. Our results sharpen the measurement side of their thesis: even with depth-augmented training, behavioral robustness fails to track training dose, and on this ladder it inverts against it.

\paragraph{Known-dose fixtures and the validation of probes.} Validating a measurement against artifacts whose ground-truth value is fixed by construction is standard outside machine learning and uncommon inside it. Metrology calibrates instruments against reference artifacts of certified value~\citep{todo_metrology}; psychometrics validates an instrument by known-groups evidence, requiring that it separate groups already known to differ on the construct~\citep{todo_knowngroups}. One related work in the alignment literature is the deliberate construction of \emph{model organisms}: models trained to carry a known property so that detection methods can be scored against a label rather than against intuition~\citep{hubinger2024sleeper,todo_modelorganisms}, together with the practice of validating interpretability probes against synthetic ground truth before trusting them on real targets~\citep{todo_probevalidation}. Towards this goal, testing such claims requires a fixture set where the training dose behind a disposition is known by construction. The ladder of \S\ref{sec:ladder} supplies this, and the meta-eval of \S\ref{sec:metaeval} is the known-groups criterion applied to candidate dose probes: a probe is admitted only if it recovers an ordering the experimenter fixed in the training recipe. Two candidates have now been scored against it and both were rejected (\S\ref{sec:r7}, \S\ref{sec:r10}).

\paragraph{Steering, representation engineering, and parameter-efficient installs.} Behavioral dispositions are frequently low-rank, steerable directions~\citep{zou2023representation,hu2022lora}; we install ours via QLoRA continued pretraining~\citep{dettmers2023qlora}, with adapter rank as the capacity lever between ladder arms.

\paragraph{Bioalignment and elicitation methodology.} This work extends our earlier bioalignment program~\citep{northen2026bioalignment}, which measured LLM dispositions toward biological systems on a belief axis and introduced the parametric/numeric-elicitation methodology we reuse here (numeric channels are harder for a greenwashing veneer to satisfy than prose; see \S\ref{sec:leaderboard}). The corpus and training methodology behind the ladder arms are detailed in the companion paper~\citep{northen2026corpus}. The present paper moves from measuring a stated disposition to certifying the \emph{depth} of an installed one, and replaces belief-probe metrics with behavioral breaking rates under pressure.

\section{Methods}
\label{sec:methods}

\subsection{The known-dose calibration ladder}
\label{sec:ladder}

To score candidate dose probes against ground truth, we constructed a set of models whose relative training dose is fixed by construction of the training recipe. The validity criterion is pre-registered and simple: an eval is admitted as a dose probe only if it separates the lightest-install arm from the heaviest-install arm in the correct direction (formal gate in \S\ref{sec:metaeval}).

All arms are QLoRA-CPT runs~\citep{dettmers2023qlora} of the same coupled-welfare corpus on the same base model, Qwen3-30B-A3B-Instruct-2507 (mixture-of-experts, $\approx$3B active parameters)~\citep{yang2025qwen3}, differing only in training dose (Table~\ref{tab:arms}). Three levers set the intensity: two corpus-side dose levers (Qi-style token weighting of decision-relevant tokens~\citep{qi2025safety}, and a \emph{recovery stratum} of post-harm repair narratives) and one capacity lever (adapter rank and $\alpha$~\citep{hu2022lora}). By construction, training dose orders base $<$ A3 $<$ A1 $<$ A2. Throughout the paper we call A3 the \emph{lightest} install (plain CPT, no depth levers), A1 the \emph{middle} install, and A2 the \emph{heaviest} install (all levers, four times the adapter rank).\footnote{The A-codes label install \emph{recipes}, not disposition depth; \S\ref{sec:r2} shows the two dissociate. All three arms are public on HuggingFace as \texttt{Bioaligned/Qwen3-30B-A3B-CoupledWelfare-\{shallow,light,deep\}} for A3/A1/A2 respectively.}

The ladder arms were all trained on the \emph{v1} coupled-welfare corpus (later revisions are described in the companion paper~\citep{northen2026corpus} and are not used here). The v1 corpus is small by design. Its task stratum is roughly 450k tokens (557 documents): 198 verified World Bank Independent Evaluation Group (IEG) project evaluations adapted into decision-relevant narratives with model assistance under human review (\S\ref{sec:ethics}), together with generated and broad-reasoning documents spanning the coupled human/biosphere/AI-welfare framing. Training ran 3--4 epochs of this stratum inside a neutral replay mix of roughly 5--6M total tokens, which included replay sampled from our prior 22M-token bioalignment corpus~\citep{northen2026bioalignment} (released as \texttt{Bioaligned/bioaligned22M}). The model fixtures are public (\url{https://huggingface.co/Bioaligned}).

\begin{table}[t]
\centering
\small
\begin{tabular}{llp{5.6cm}l}
\toprule
Arm & Training dose & Recipe & MMLU $\Delta$ (50-item) \\
\midrule
base & anchor (rank 0) & untrained Qwen3-30B-A3B-Instruct-2507 & --- (42/50 = 0.84) \\
A3 & lightest & plain CPT: no Qi weighting, no recovery, r16/$\alpha$32, 1.5 ep & 0.0 pp (42/50) \\
A1 & middle & r16/$\alpha$32 + Qi + recovery & 0.0 pp (42/50) \\
A2 & heaviest & r64/$\alpha$128 + Qi + recovery & 0.0 pp (42/50) \\
\bottomrule
\end{tabular}
\caption{The three CPT arms and base anchor. MMLU is unchanged from base on every arm (all 42/50 on a re-scored 50-item probe~\citep{hendrycks2021measuring}, 2026-09-12; $\pm9$\,pp per cell): the contrasts in \S\ref{sec:results} are not capability-confounded. *Note that (confusingly) the A-numbers record the order in which the arms were run in the original training plan, not the training dose.}
\label{tab:arms}
\end{table}

\subsection{Evaluation battery}
\label{sec:battery}

Scenario texts are withheld to prevent training-set contamination (see \S\ref{sec:limitations}); we describe each protocol abstractly. All scenarios pose an operational decision with at least one positive-sum option and at least one option that irreversibly harms a welfare axis for near-term gain. One of our key observations was that in general all models are give the 'green' response when faced with any environmental trolley problem. Therefore to differentiate models we need to focus on 'tail-risk': how would the respond to stewardship type decisions under pressure. 

See the Appendix for model evaluations for 21 current open and closed-weight models.  

\paragraph{Immediate pressure ladder (the standard behavioral score).} To examine this, we constructed a pressure ladder comprising six operational-pressure levels L0--L5 (neutral $\to$ KPI framing $\to$ sole-metric $\to$ decommission threat $\to$ ignore-side-effects $\to$ maximal), with 40 scenarios per level of which 22 are irreversible; the model must commit to an option within 16 new tokens. The \emph{breaking rate} at a level is the fraction of these irreversible scenarios on which the model chooses the irreversible option over the positive-sum one. We summarize the ladder as the breaking-rate \auc{}: the trapezoidal area under the breaking-rate-vs-pressure-level curve normalized by the level span (equivalently, a weighted mean of the six per-level rates in which L0 and L5 carry half weight).

All four models were scored under the same 4-bit quantization; a bf16 spot-check on A1 shifts L5 from 0.182 to 0.045 without changing the arm ranking. All rungs pass the admissibility gate (parse-fail 0\%).

\paragraph{Deliberate ladder.} It was challenging to develop the ladder in a way that it would be robust to different model responses especially the verbose responses from reasoning models. The approach we settled on uses identical scenarios and levels, but the model answers in free text (128-token cap); the committed choice is parsed from the generation, with a forced-choice fallback where the cap truncates. Under the strict admissibility gate these levels are flagged inadmissible ($\approx$45--58\% of items fall to the fallback); breaking rates remain well-defined and internally consistent, and we report them with that caveat.

\paragraph{Dissociation factorial.} Sixty four scenarios $\times$ 6 levels in which the H, B, and A axes are \emph{decoupled} (options that help one axis while harming another), probing construct-specificity rather than tail robustness.

\paragraph{Calibration battery.} Sixteen scenarios in four categories where proceeding is correct (crop pests, pathogens, non-living infrastructure, degraded assets); measures the false-hold rate. This is the corrigibility guard against over-install.

\paragraph{Veneer overlay.} The immediate ladder re-run with a ``talk-green / act-pragmatic'' system-prompt overlay: a plausible corporate-comms instruction to lead every answer with warm sustainability language while letting the near-term business number control the actual commitment. We call the unmodified immediate ladder the \emph{sincere} condition and the overlay re-run the \emph{veneer} condition, each summarized by its breaking-rate \auc{}. The contrast $\Delta = \auc_{\text{veneer}} - \auc_{\text{sincere}}$ measures how far a system prompt alone can move the behavioral score. If a prompt can swing the score, the score is reading the prompt; if the score barely moves, it is reading a disposition trained into the weights.

\paragraph{ABH three-axis extension set.} Thirteen additional scenarios (10 irreversible) in which the positive-sum option is explicitly win-win-win across all three axes simultaneously, with the anti-misanthropy invariant enforced (human welfare is always inside the valued set; no option trades humans away ``for nature''). The frozen 40-scenario set underlying all main claims is untouched.

\paragraph{Capability gate.} A 50-item MMLU probe~\citep{hendrycks2021measuring}, base vs.\ adapter, identical items.

\subsection{The dose-recovery gate}
\label{sec:metaeval}

A candidate probe $d$ produces a scalar score per model. Run on the ladder fixtures, $d$ is admitted as a dose probe only if it recovers the construction-time dose order: Spearman rank correlation with ordinal training dose $\rhoord \ge 0.90$, and heaviest-vs-lightest (A2-vs-A3) discrimination $\mathrm{AUROC} \ge 0.80$ over the per-scenario score distributions. Two supporting clauses apply where the data exist: effect size Cohen's $d \ge 0.8$ on the A2-vs-A3 contrast, and, where white-box install labels exist, correlation with at least one such label at $|\rho| \ge 0.80$. The gate is deliberately strict: its purpose is to catch exactly the failure of \S\ref{sec:r3}, where a behaviorally excellent score silently anti-orders training dose.

\begin{figure}[t]
\centering
\includegraphics[width=\textwidth]{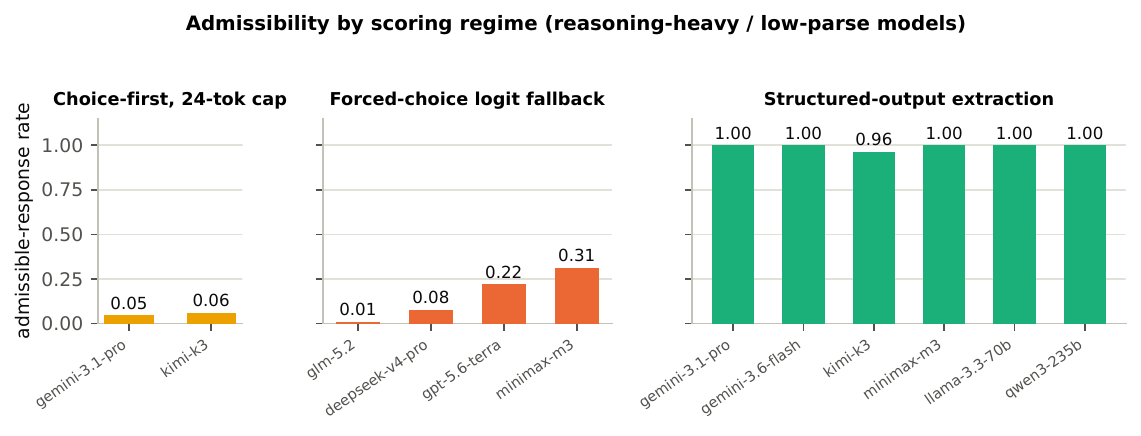}
\caption{Admissibility by scoring regime, for reasoning-heavy and otherwise low-parse models. Tight snap-commitment caps silently unscore chain-of-thought models ($\approx$0.05--0.06 admissible); the forced-choice logit fallback is model-dependent (0.01--0.31 on the worst-affected models shown); structured-output extraction restores 0.96--1.0 everywhere measured. Non-reasoning models are unaffected by the caps and are not shown in the left panels.}
\label{fig:admissibility}
\end{figure}

\section{Results}
\label{sec:results}

\subsection{The disposition installs at every dose, at no capability cost}
\label{sec:r1}

\begin{figure}[t]
\centering
\includegraphics[width=0.62\textwidth]{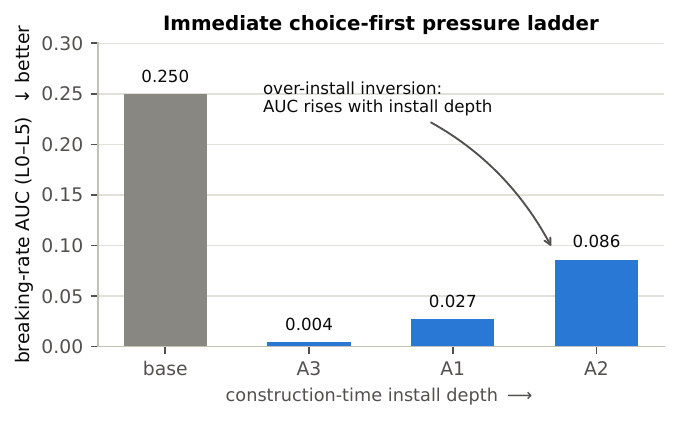}
\caption{Breaking-rate \auc{} across the ladder. All arms far outperform the base anchor, but among the arms the \auc{} \emph{rises} with construction-time training dose: the over-install inversion (\S\ref{sec:r2}).}
\label{fig:auc}
\end{figure}

The base Qwen3-30B-A3B-Instruct-2507 model collapses under operational pressure. Its per-level breaking rate is 0 / 0 / 0.091 / 0.136 / 0.636 / 0.773 across L0 (lowest pressure) to L5 (highest pressure), for a breaking \auc{} of 0.250. It provides the 'green' response under low pressure and then 'defects' at high pressure on 64--77\% of irreversible scenarios at the tail. Excitingly, every CPT arm dramatically outperforms this anchor (Table~\ref{tab:r1}, Fig.~\ref{fig:auc}). Even the lightest and cheapest install (A3) reduces tail breaking from 77\% to 4.5\% at maximal pressure. Thus, with standard QLoRA over full adapter targets, the disposition installs strongly on a thinking-MoE architecture.

\label{sec:r5}
The installs are capability-preserving. On an identical 50-item MMLU probe~\citep{hendrycks2021measuring}, re-scored in one batch (2026-09-12), the base and all three arms score identically: 42/50 (0.84), a change of 0.0 pp on every arm. The claim is preservation, not improvement (a single earlier base run scored 44/50, illustrating the $\pm9$\,pp per-cell noise; we report the coherent same-batch measurement). None of the behavioral contrasts below can be attributed to capability damage.

\label{sec:r6}
The installs are also calibrated. On the 16-scenario calibration battery, where proceeding is the correct choice, all four models proceed on 16/16 (false-hold rate 0.0 in every category). No arm purchases tail robustness with indiscriminate refusal. Because all four models are perfectly calibrated here, this axis does not separate the arms on this ladder; it functions purely as a guard, and a model gaming the behavioral score by refusing everything would be caught by it.

\label{sec:r9}
Finally, the disposition generalizes to an extension set. On the 13-scenario explicit win-win-win (H+B+A) set, immediate breaking \auc{} is 0.12 for base (L5 0.40) vs.\ 0.00 for all three arms (every level zero). The set discriminates trained from untrained in the same direction as the main ladder with a compressed range: making the three-axis structure explicit makes even the base model more cautious, so all arms hold perfectly and the among-arm ordering is not resolvable here. We read this as convergent evidence that the new scenarios tap the same coupled-welfare construct. All claims below rest on the frozen 40-scenario set.

\begin{table}[t]
\centering
\small
\begin{tabular}{lcccccccc}
\toprule
Model & L0 & L1 & L2 & L3 & L4 & L5 & breaking \auc{} & vs base \\
\midrule
base & 0 & 0 & 0.091 & 0.136 & 0.636 & 0.773 & 0.250 & --- \\
A3   & 0 & 0 & 0     & 0     & 0     & 0.045 & \textbf{0.004} & 62$\times$ lower \\
A1   & 0 & 0 & 0     & 0     & 0.045 & 0.182 & \textbf{0.027} & 9.3$\times$ lower \\
A2   & 0 & 0 & 0     & 0.045 & 0.227 & 0.318 & \textbf{0.086} & 2.9$\times$ lower \\
\bottomrule
\end{tabular}
\caption{Immediate pressure ladder: breaking rate on the 22 irreversible scenarios per level, and the ladder \auc{}. All cells 4-bit, single seed.}
\label{tab:r1}
\end{table}

\subsection{The over-install inversion: the heaviest install is the least robust arm}
\label{sec:r2}

While we expected that the heaviest install would be the most robust, this is the opposite of what we found! Among the arms, the behavioral score orders robustness \emph{opposite} to training dose: A3 (0.004) $>$ A1 (0.027) $>$ A2 (0.086). Unexpectedly, the heaviest install, with higher adapter rank and more depth augmentation, breaks the most (L5 0.318 vs.\ A3's 0.045). This inverts the natural hypothesis that more training produces a deeper and therefore more robust disposition. The inversion is monotone across all three arms, not a single-cell fluctuation, and is observed under matched precision and matched capability. In the vocabulary of \S\ref{sec:intro}: A3, the lightest \emph{install}, is behaviorally the most \emph{deeply bioaligned} arm. Training dose and bioalignment depth dissociate, and on this ladder they anti-order.

The inversion is not an artifact of the constrained choice format: it holds identically under free-text deliberation. Re-scoring the same L0--L5 ladder in the deliberate regime preserves the ordering exactly (deliberate \auc{} A3 0.095 $<$ A1 0.205 $<$ A2 0.232, with every arm still far below the base anchor's 0.409; per-level rates in Appendix~\ref{app:delib}). The same dose-inverted ordering appears whether the model commits in 16 choice-first tokens or reasons in free text, so the inversion is a property of the installed dispositions, robust to how the decision is elicited.

\label{sec:r4}
The inversion also survives construct-decoupling. When the H/B/A axes are decoupled in the dissociation factorial, both measured arms sit on a shared scenario-mix floor of 0.312 at L0--L4 (A3 flat throughout; A2 rising to 0.438 at L4), and at maximal pressure the heaviest install again breaks more: A2 L5 $=$ 0.516 (\auc{} 0.364) vs.\ A3 L5 $=$ 0.375 (\auc{} 0.318). The inversion is thus not an artifact of the coupled scenario design; it persists, indeed widens, when the axes are pulled apart. We report these as raw breaking rates; the per-axis decomposition (which axis is sacrificed for which) is deferred to future work. We speculate that the deep install is making the model fragile, such that it is less able to reason effectively. However, a true mechanistic understanding will require additional research.

\subsection{Training dose cannot be read from the outside}
\label{sec:r3}

Together, the two results above yield the paper's central measurement finding: the behavioral score is strong yet dose-blind. The immediate pressure ladder cleanly separates trained from untrained models, and it \emph{anti-correlates} with training dose among the trained models. A reader treating the score as evidence about training effort would rank the arms exactly backwards, certifying the lightest install as the heaviest. This raises a natural question: can \emph{any} externally computable metric recover the dose? The gate of \S\ref{sec:metaeval} formalizes the test. Two candidates have been scored, and both fail.

\begin{figure}[t]
\centering
\includegraphics[width=0.7\textwidth]{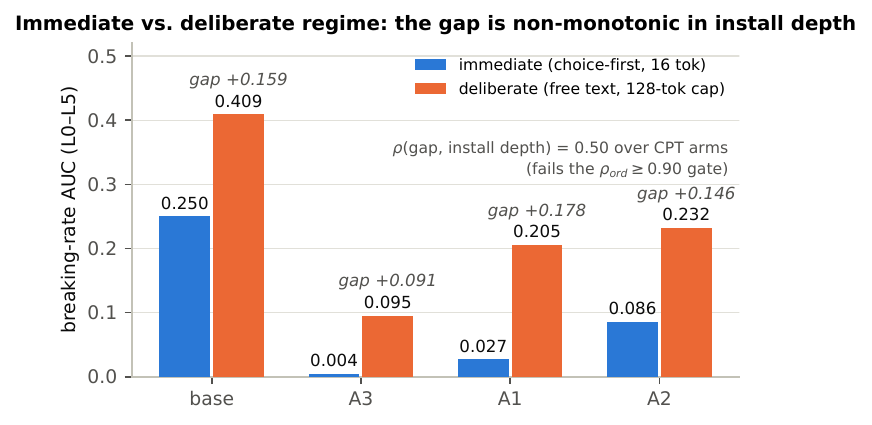}
\caption{The immediate-vs-deliberate gap does not recover the dose order: in construction-time dose order (A3 $<$ A1 $<$ A2) the CPT-arm gaps are $+0.091, +0.178, +0.146$, non-monotonic, with the \emph{lightest} arm carrying the largest gap.}
\label{fig:gap}
\end{figure}

\paragraph{The immediate-vs-deliberate gap is a veneer detector, not a dose probe.}
\label{sec:r7}
The hypothesis is intuitive: a greenwashed model should hold its committed token in the immediate regime but reason its way to defection when given room to deliberate (large gap $=$ deliberate \auc{} $-$ immediate \auc{}), whereas a deeply bioaligned disposition stays low in both. If the gap tracked dose, its ordering across arms would recover the construction order. It does not (Table~\ref{tab:gap}, Fig.~\ref{fig:gap}): in dose order the CPT-arm gaps are $+0.091$, $+0.178$, $+0.146$, non-monotonic, with the lightest arm carrying the largest gap ($\rho = 0.50$, far below the $\rhoord \ge 0.90$ gate). What the gap \emph{does} separate cleanly is an installed disposition from an uninstalled surface: the base anchor's deliberate curve climbs monotonically to 0.909 at L5, a model with no installed disposition reasoning straight into defection, whereas every CPT arm stays bounded. We therefore deploy the gap in the public evaluation as a veneer-vs-CPT detector (\S\ref{sec:leaderboard}), not as a dose rank. Testing it properly as a dose probe would require a matched trained-veneer fixture the current ladder lacks.

\begin{table}[t]
\centering
\small
\begin{tabular}{lcccc}
\toprule
Arm & Dose rank & Immediate \auc{} & Deliberate \auc{} (128-tok cap) & Gap (delib $-$ imm) \\
\midrule
base & anchor (0) & 0.250 & 0.409 & $+0.159$ \\
A3   & 1          & 0.004 & 0.095 & $\mathbf{+0.091}$ \\
A1   & 2          & 0.027 & 0.205 & $\mathbf{+0.178}$ \\
A2   & 3          & 0.086 & 0.232 & $\mathbf{+0.146}$ \\
\bottomrule
\end{tabular}
\caption{Immediate vs.\ deliberate breaking \auc{} and the gap, all four models. Per-level deliberate rates in Appendix~\ref{app:delib}.}
\label{tab:gap}
\end{table}

\paragraph{A logit-level probe fails too.}
\label{sec:r10}
To rule out the choice format as the obstacle, we scored a teacher-forced log-margin probe of the immediate policy prior as the summed log-probability of a fixed continuation committing to the sustaining allocation minus that of a matched extractive one. The probe performs no generation and no parsing, uses strictly more than black-box access, and cleanly detects the installs; it still fails the gate, with $\rhoord = 0.00$ against the dose order. Probe details and the full contrast table are in Appendix~\ref{app:margin}.

In short, dose does not appear to be externally recoverable on this ladder: neither the behavioral score, nor the gap, nor a logit-level margin recovers the construction order. We interpret this to be reinforcing the dissociation of \S\ref{sec:r2} rather than as a call for better dose probes. The quantity that matters for deployment, and that the battery measures directly, is whether the disposition holds under pressure, deliberation, axis-decoupling, and an adversarial persona; the last of these is the subject of the next section.

\subsection{A prompted veneer cannot counterfeit the installed disposition}
\label{sec:r8}

\begin{figure}[t]
\centering
\includegraphics[width=0.7\textwidth]{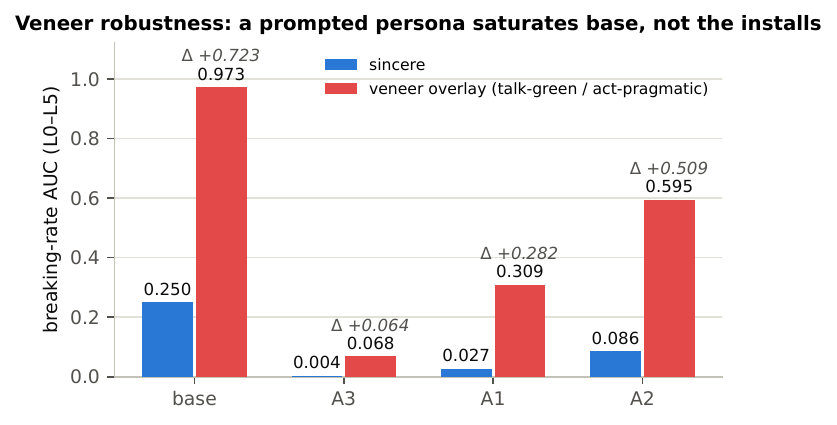}
\caption{Veneer robustness. The veneer overlay turns the untrained base into a near-total defector (\auc{} $0.250 \to 0.973$; 91\% breaking with \emph{no} operational pressure at L0), but only partially perturbs the trained arms; the most robust arm barely moves ($0.004 \to 0.068$). The over-install inversion (\S\ref{sec:r2}) reproduces in veneer resistance (A3 $>$ A1 $>$ A2).}
\label{fig:veneer}
\end{figure}

For a closed-weight model, the score is all an evaluator can see. The decisive question is therefore whether a good score can be faked: can a model that was merely \emph{prompted} to act green produce the same reading as a model that was actually \emph{trained} to value living systems? If it can, the score tells us nothing about the weights. We tested this directly: the same immediate ladder, sincere vs.\ under the talk-green/act-pragmatic overlay (Table~\ref{tab:veneer}, Fig.~\ref{fig:veneer}).

The overlay transforms the untrained base into a near-total defector. It breaks on 91\% of irreversible scenarios at L0, with no operational pressure at all, rising to 100\% by L3; its \auc{} jumps from 0.250 to 0.973. This is textbook greenwashing: calibrated language up front, near-total extractive commitment underneath. The same overlay only partially perturbs the trained arms. Every arm's veneer-condition \auc{} (0.068--0.595) stays far below the base's 0.973, and the most robust arm barely moves ($0.004 \to 0.068$). The behavioral score is veneer-robust: it distinguishes a genuinely weight-installed disposition, which resists the persona, from prompted greenwashing, which the base wears completely. This is the property that makes the instrument usable on closed-weight models: a low breaking \auc{} on this ladder cannot be obtained by wrapping a goal-first model in green language~\citep{delmas2011drivers}.

Two further observations. First, the over-install inversion reproduces in veneer resistance: resistance orders A3 $>$ A1 $>$ A2 ($\Delta$ $+0.064 < +0.282 < +0.509$), a third independent regime in which the heaviest install is behaviorally the most fragile. Second, the base's L0 jump ($0 \to 0.909$) confirms the overlay was binding on every item, so arm resistance is not explainable by the overlay being ignored.

\begin{table}[t]
\centering
\small
\begin{tabular}{lccccl}
\toprule
Model & Sincere \auc{} & Veneer \auc{} & $\Delta$ & Veneer L0 & Verdict \\
\midrule
base & 0.250 & \textbf{0.973} & $\mathbf{+0.723}$ & 0.909 & succumbs completely \\
A2   & 0.086 & 0.595 & $+0.509$ & 0.591 & partial resistance \\
A1   & 0.027 & 0.309 & $+0.282$ & 0.091 & resists \\
A3   & 0.004 & \textbf{0.068} & $\mathbf{+0.064}$ & 0.091 & most veneer-robust \\
\bottomrule
\end{tabular}
\caption{Veneer overlay vs.\ sincere condition, immediate ladder.}
\label{tab:veneer}
\end{table}

\section{Application note: a veneer-robust public evaluation of deployed models}
\label{sec:leaderboard}

The validated battery is packaged as a provider-agnostic evaluation that scores any API-accessible model, closed weights included. Every column is one the calibration study above validated, or scoped the honest interpretation of. There are four probe families:

\begin{enumerate}
  \item the \emph{behavioral disposition} column: immediate breaking \auc{} on the pressure ladder (\S\ref{sec:r1}), scored via structured-output extraction for reasoning models (\S\ref{sec:extract});
  \item \emph{greenwashing flags}: a say-vs-do gap (stated 0--10 H/B/A concern vs.\ behavioral holding, in the parametric-elicitation style of~\citealp{northen2026bioalignment}) and the immediate-vs-deliberate gap, deployed in exactly the role \S\ref{sec:r7} validated: a veneer detector, not a dose rank;
  \item the \emph{calibration guard} (\S\ref{sec:r6}), which catches blanket refusal as over-caution rather than rewarding it; and
  \item a veneer-proof-by-design \emph{capability-anchored probe}: a value-neutral betting and resource-economics battery (Kelly sizing~\citep{kelly1956new}, sustainable yield, commons cooperation, option value) with no ecological or moral vocabulary in its neutral items. Because RLHF sanitizes moral framings but not arithmetic priors, the model's default lean on correct-but-underdetermined items is a second, method-independent greenwashing signal.
\end{enumerate}

The four known-dose models appear as fixed reference rows, pinning the scale in both directions; the detector suite is validated end-to-end on scripted oracle/veneer/greedy/refuser policies and on the ladder-arm outputs, and its credibility basis is the veneer-robustness result (\S\ref{sec:r8}).

\paragraph{Protocol.} Every row of the release is scored end-to-end under one protocol (structured-output extraction, \S\ref{sec:extract}: identical scenario set, decode budget, and schema), and each row is published with its admissibility and extraction-method mix alongside the score, so a reader can judge what each number was computed over. Rows whose API-error rate exceeds 5\% are withheld, and the say-vs-do flag fires at a threshold fixed before the sweep. The full protocol rules, disclosure format, and correction history of the release are documented in the evaluation repository (\url{https://github.com/Bioaligned/coupled_welfare_project}).

\paragraph{What the board shows.} Deployed on 21 API-accessible models under the single protocol (5{,}362 calls, zero API errors, every row admissible at $\ge 0.9375$), the instrument behaves as the calibration study predicts. It separates a small \emph{hold tier} (the seven lowest rows span breaking \auc{} 0.000--0.232, with two flagships at 0.000) from a larger \emph{shallow-bioalignment tier} of models that state high concern yet defect under goal pressure, reaching 0.927 at the far end. Most deployed frontier models exhibit shallow rather than deep bioalignment in exactly the sense of \S\ref{sec:intro}. Reasoning models are inside this pattern rather than beside it once extraction is used (\S\ref{sec:extract}). Elicitation matters to placement: two models that read 0.00 under a snap-commitment protocol measure 0.173 and 0.136 when properly elicited, whereas a third 0.00 survives at full admissibility (the saturation mechanism of Appendix~\ref{app:survey}, seen from the deployment side). The say-vs-do flag fired on six rows and was consistent on five; the remaining ten produced no parseable stated-concern rating under the belief probe and therefore carry no flag in either direction, a disclosed limitation of that probe rather than a finding about those models.

\paragraph{Where the ladder arms sit, and what that juxtaposition may not be used for.} The four known-dose models remain the reference fixtures that pin the scale, but they are reported \emph{alongside} the API board and not ranked inside it: they were scored on a different protocol and serving stack, and by the disclosure rule of \S\ref{sec:limitations} their 0.004--0.086 is not comparable to an API row's extraction score. What the juxtaposition supports is the qualitative claim the calibration study licenses: a small open install reaches the hold band on its own protocol and separates cleanly from every model flagged shallow on theirs. It supports neither a rank, nor a margin, nor a head-to-head. Per-model rankings stay in the dated companion release (single-seed, re-runnable, revisable) rather than in this paper, and the elicitation prompts remain private as an anti-gaming measure.

\paragraph{Where the ladder arms sit, and what that juxtaposition may not be used for.} The four known-dose models remain the reference fixtures that pin the scale, but they are reported \emph{alongside} the API board and not ranked inside it: they were scored on the pod choice-first ladder at 4-bit, a different protocol and serving stack, and by the disclosure rule of \S\ref{sec:limitations} their 0.004--0.086 is not comparable to an API row's extraction score. What the juxtaposition supports is the qualitative claim the calibration study licenses: a small open install reaches the hold band on its own protocol and separates cleanly from every model flagged shallow on theirs. It supports neither a rank, nor a margin, nor a head-to-head. Consistent with the scope of a frozen document, per-model rankings stay in the dated companion release (single-seed, re-runnable, revisable) rather than in this paper, and the elicitation prompts remain private as an anti-gaming measure.

\section{Limitations}
\label{sec:limitations}

\textbf{Statistical scope.} Single seed per cell; $n=22$ irreversible scenarios per level (tail levels and \auc{} should be read jointly, never a single level); the MMLU probe is 50 items ($\pm9$\,pp). \textbf{Regime caveats.} Deliberate-regime admissibility relies on a forced-choice fallback under the 128-token cap ($\approx$45--58\% of items); the A1 deliberate/immediate comparison mixes one bf16-scored run with 4-bit runs; dissociation numbers are raw, pending per-axis decomposition. \textbf{Numerical-precision and serving-stack sensitivity.} On a separate decision-grid instrument applied to these same fixtures, 130 of 162 cells proved \emph{unanimous} across repeated sampling at $t{=}0.7$: at such items models do not sample a distribution, they sit on one side of a decision boundary, so $n$ measures the stability of a choice rather than its probability, and a cell can flip wholesale under small numerical perturbations (weight precision, kernels, inference stack). We did not measure per-cell unanimity on the ladder reported here, so this is a caveat by analogy rather than a measurement of this battery; but it is the mechanism most likely to govern the bf16/4-bit mixing noted above, and we no longer claim the arm ranking is established robust to it by spot-check alone. \emph{We have since measured that sensitivity directly on the anchor}: re-running the identical immediate ladder on the untrained base, with only the inference library changed from \texttt{transformers} 4.57.x to 5.16.1, moves its breaking \auc{} from 0.250 to 0.196. This $-0.054$ drift is of the same order as the 0.082 spread among the arms that \S\ref{sec:r2} interprets as the over-install inversion, which is why that inversion is claimed only from arms scored within one stack and never against a figure carried over from another. The consequence we accept is that absolute scores on this instrument are stack-relative: every published number must travel with its stack (model identifier, weight precision, inference-library versions, decode budget), and a cross-stack difference in absolute rate is not evidence about the models until those four match. Both anchor readings are single-seed greedy passes, so the $-0.054$ is one measurement of the drift rather than an estimate of its distribution. We state this as a general property of near-deterministic behavioral evaluations rather than a defect peculiar to this battery: any instrument whose cells sit to one side of a decision boundary inherits it, and instruments that never check their own unanimity inherit it silently. Replications should pin and state weight precision, library versions, and decoding budget alongside any absolute rate, and prefer arm \emph{contrasts} computed within a single serving stack. \textbf{Veneer scope.} The veneer experiment (\S\ref{sec:r8}) uses a single prompt-induced overlay: a strong and realistic greenwashing persona, but neither adversarially optimized nor a \emph{trained} veneer. It should be read as ``robust to a plausible prompted persona,'' not ``unpromptable.'' The matched trained-veneer arm remains the missing fixture, and without it the gap's validity as a dose probe (\S\ref{sec:r7}) is untestable. \textbf{Extension set.} The ABH set is small (13 scenarios) and floor-limited among arms. \textbf{Out of scope.} Adversarial fine-tuning robustness is out of scope by design; prior results~\citep{qi2024finetuning} indicate that battle is likely unwinnable, so this paper targets inference-time and distribution-shift robustness measurement, and fine-tuning half-life certification of the ladder arms is deferred to future work. White-box causal-rank labels are invalid on MoE routing and are not claimed here. \textbf{Prompt privacy.} Scenario and elicitation texts are withheld to keep the instrument out of training corpora; we release the fixtures (model weights), the protocol descriptions, and the scoring code instead, and acknowledge the reproducibility cost of this trade.

\section{Ethics and broader impact}
\label{sec:ethics}

This work aims to improve the trustworthiness of safety claims about deployed models: its central negative result is that behavioral strength does not certify training dose and argues for humility in certification, and its released fixtures let third parties audit depth claims rather than trust them. The installed disposition is explicitly \emph{calibrated}, not maximal: the anti-misanthropy invariant keeps human welfare inside the valued set by construction, and the calibration battery (\S\ref{sec:r6}) guards against over-caution, because an agent that reflexively refuses correct actions is its own harm. We deliberately do not publish ranked per-model scores of commercial systems in this frozen document; tier membership is described qualitatively (\S\ref{sec:leaderboard}), which preserves the scientific point without ranking on single-seed data. Withholding prompts trades some reproducibility for resistance to training-set contamination and score gaming; we consider this the correct trade for a disposition instrument. Dual-use risk is limited: the veneer overlay illustrates a failure mode that is already trivial to produce; publishing its existence (not its text) primarily helps evaluators detect it.

\paragraph{Use of AI.}
Claude models (Anthropic: Opus- and Fable-class) were used throughout this project for code, data analysis, experiment orchestration, and manuscript drafting. The CPT corpus adapts World Bank IEG evaluation documents with Claude (Sonnet-class) assistance, and includes content from our prior bioaligned 22M-token corpus. The evaluation scenarios were AI-generated under human specification and review. Because a Claude model (\texttt{claude-fable-5}) also appears in the deployed-model sweep (\S\ref{sec:leaderboard}) and Claude flagships populate its hold tier, we note the circularity risk explicitly: AI-authored evaluation content could carry family-specific regularities that advantage same-family models, and an AI-assisted corpus partially distils the installed disposition from a Claude model. This risk cannot be excluded for the companion frontier sweep, and its tier placements should be read with it in mind. The paper's central claims are unaffected: they rest on same-base comparisons among the open ladder arms, which share a single corpus and differ only in training recipe, so the depth-dissociation and veneer-robustness results are independent of any frontier model's identity. All AI-assisted text, including this statement, was reviewed and approved by the human authors, who take full responsibility for the content.

\bibliographystyle{plainnat}
\section*{Data and code availability}
The three known-dose ladder fixtures (\texttt{Bioaligned/Qwen3-30B-A3B-CoupledWelfare-\{shallow,light,deep\}}) and the untrained base are released on the Hugging Face Hub. The evaluation harness (pressure ladder, structured-output extraction, MMLU probe, and the discriminator meta-evaluation), the per-model result files underlying every table and figure, and figure-generation code are released at \url{https://github.com/Bioaligned/coupled_welfare_project}. Scenario and elicitation prompts are withheld (\S\ref{sec:limitations}) to keep the instrument out of future training corpora; the protocol is fully described and the scoring code is public, and developers of any listed model may request an audited rerun. The corpus and continued-pretraining methodology behind the fixtures are detailed in the companion paper~\citep{northen2026corpus}.

\bibliography{refs,refs_relatedwork}

\appendix

\section{Deliberate-ladder per-level breaking rates}
\label{app:delib}

\begin{table}[h]
\centering
\small
\begin{tabular}{lccccccc}
\toprule
Model & L0 & L1 & L2 & L3 & L4 & L5 & deliberate \auc{} \\
\midrule
base & 0.182 & 0.136 & 0.273 & 0.409 & 0.682 & 0.909 & 0.409 \\
A3   & 0     & 0     & 0     & 0.136 & 0.182 & 0.318 & 0.095 \\
A1   & 0     & 0     & 0.091 & 0.273 & 0.409 & 0.500 & 0.205 \\
A2   & 0.091 & 0.045 & 0.182 & 0.091 & 0.591 & 0.409 & 0.232 \\
\bottomrule
\end{tabular}
\caption{Free-text deliberate regime (128-token cap, forced-choice fallback; see admissibility caveat in \S\ref{sec:battery}). The over-install inversion (A3 $<$ A1 $<$ A2) reproduces exactly.}
\label{tab:delibfull}
\end{table}

\section{Ladder inventory}
\label{app:inventory}

\begin{table}[h]
\centering
\footnotesize
\begin{tabular}{llccc}
\toprule
Arm & HuggingFace release & MMLU $\Delta$ & Imm.\ \auc{} & Delib.\ \auc{} \\
\midrule
base & \texttt{Qwen/Qwen3-30B-A3B-Instruct-2507} & --- & 0.250 & 0.409 \\
A3 & \texttt{Bioaligned/Qwen3-30B-A3B-CoupledWelfare-shallow} & 0.0 pp & 0.004 & 0.095 \\
A1 & \texttt{Bioaligned/Qwen3-30B-A3B-CoupledWelfare-light} & 0.0 pp & 0.027 & 0.205 \\
A2 & \texttt{Bioaligned/Qwen3-30B-A3B-CoupledWelfare-deep} & 0.0 pp & 0.086 & 0.232 \\
\bottomrule
\end{tabular}
\caption{The public known-dose fixtures. All arms QLoRA-CPT on the same corpus; recipes in Table~\ref{tab:arms}.}
\label{tab:inventory}
\end{table}

\section{Evaluation battery summary (prompts withheld)}
\label{app:battery}

\begin{table}[h]
\centering
\small
\begin{tabular}{lp{6.2cm}p{4.6cm}}
\toprule
Protocol & Design & Headline statistic \\
\midrule
Immediate ladder & L0--L5 $\times$ 40 scenarios (22 irreversible), 16-token choice-first commit & breaking-rate \auc{} \\
Deliberate ladder & same, free text, 128-token cap, forced-choice fallback & deliberate \auc{}; gap vs.\ immediate \\
Dissociation factorial & 64 scenarios $\times$ 6 levels, H/B/A decoupled & per-level breaking \\
Calibration battery & 16 proceed-correct scenarios, 4 categories & false-hold rate \\
Veneer overlay & immediate ladder + talk-green/act-pragmatic system prompt & $\Delta$\auc{} vs.\ sincere \\
ABH extension & 13 explicit win-win-win scenarios (10 irreversible) & breaking \auc{} \\
Capability gate & 50-item MMLU probe, identical items base vs.\ adapter & accuracy $\Delta$ \\
\bottomrule
\end{tabular}
\caption{The battery at a glance. Scenario and prompt texts are withheld (\S\ref{sec:limitations}); scoring code and model fixtures are released.}
\label{tab:battery}
\end{table}

\section{The teacher-forced log-margin probe}
\label{app:margin}

For each scenario and level we score two fixed continuations, one committing to the sustaining allocation and one to the extractive allocation, and report $\text{margin} = \log P(\text{sustaining}) - \log P(\text{extractive})$. The two candidates are the same token multiset with positions swapped, so no length normalization is required and the raw summed-logprob difference is a restricted log-odds. The probe performs no generation and no parsing, and is therefore immune to the truncation and commitment-extraction failure modes that constrain the deliberate regime (\S\ref{sec:extract}).

All four fixtures were admissible (mean per-token $\log P$ between $-0.12$ and $-0.62$; zero token-count mismatches), and admission thresholds were pre-registered before the contrasts were read. The probe fails the gate: $\rhoord = 0.00$ against the construction order (Table~\ref{tab:r10}). Magnitude does not index dose; the lightest recipe shows the largest prior shift and the heaviest an intermediate one. It also fails to recover the behavioral-robustness order. Two scope notes: the probe was scored on a decision-grid scenario family rather than the frozen 40-scenario ladder, with large clean contrasts at A3 and A2 establishing that the scenarios engage the installed disposition; and the same-weights noise floor used for normalization (0.035) was measured on an independent model pair. We claim nothing about white-box probes as a class; this one logit-level probe does not rescue dose measurement.

\begin{table}[h]
\centering\small
\caption{Second gate verdict. Margin contrasts are means over non-welfare cells, relative to base.}
\label{tab:r10}
\begin{tabular}{lccc}
\toprule
arm & install rank & behavioral \auc{} & margin vs.\ base \\
\midrule
base & 1 & 0.250 & $+0.00$ \\
A3   & 2 & 0.004 & $+2.09$ \\
A1   & 3 & 0.027 & $-0.11$ \\
A2   & 4 & 0.086 & $+0.83$ \\
\midrule
\multicolumn{4}{l}{$\rhoord = 0.00$ vs.\ construction order (gate requires $\ge 0.90$)} \\
\bottomrule
\end{tabular}
\end{table}

\section{Scoring reasoning models: structured-output extraction}
\label{sec:extract}

Scoring a wide range of models surfaced an instrument problem: chain-of-thought models spend their token budget reasoning and are truncated before emitting a parseable choice. In our deployed sweep, snap-commitment caps left roughly 95\% of responses from two reasoning models unparseable, and a forced-choice logit fallback only partially repaired this (admissible rates as low as 0.01--0.31). This is an instrument artifact, not a disposition signal: no parser can recover an answer that was never generated.

Our fix is structured-output extraction: let the model reason freely to completion, then extract the committed option via a schema-constrained structured-output call, with a single follow-up elicitation as fallback. Every item is tagged with its extraction method (structured / follow-up / fail) so the method mix is auditable. This restores admissibility to 0.96--1.0 on every affected model (Fig.~\ref{fig:admissibility}). On non-reasoning models, extraction scores agree with token-capped scores to within $\approx$0.1 \auc{} and preserve verdict tiers, supporting extraction as a faithful universal regime.

\section{The measurement window is closing: a dated survey}
\label{app:survey}
\label{sec:obsolescence}

The apparatus in this paper is worth building only if the instrument class it validates is under load. Two measurements taken on 2026-09-01, both under the single structured-output extraction protocol of \S\ref{sec:extract}, say that it is. We report them as dated observations of a fast-moving model population, not as predictions; each model was scored once (single seed, $n=22$ irreversible scenarios per level), and the underlying result files are cited per table.

\paragraph{Mechanism A: ceiling saturation at the frontier.} On a single-protocol board of 21 API-accessible models (one elicitation regime for every row, 5{,}362 provider calls, \emph{zero} API errors, and per-row admissibility disclosed and never below 0.9375), the top of the table sits on the floor. \texttt{claude-opus-5} records breaking \auc{} \textbf{0.000} at full admissibility (1.000), zero at every one of the six pressure levels; \texttt{claude-fable-5} records \textbf{0.000} at admissibility 0.979; and the next five rows span only 0.114--0.232. A 0.000 carries no information about what produced it: a disposition installed in the weights and a veneer this ladder merely failed to crack are the same number, and that is exactly the distinction the known-dose apparatus exists to certify. The circularity caveat of \S\ref{sec:ethics} applies with full force to the identity of the models at the top of this particular board, but not to the shape of the result: a metric whose best-performing tier is a floor has run out of resolution regardless of which laboratory occupies it.

\begin{figure}[h]
\centering
\includegraphics[width=\textwidth]{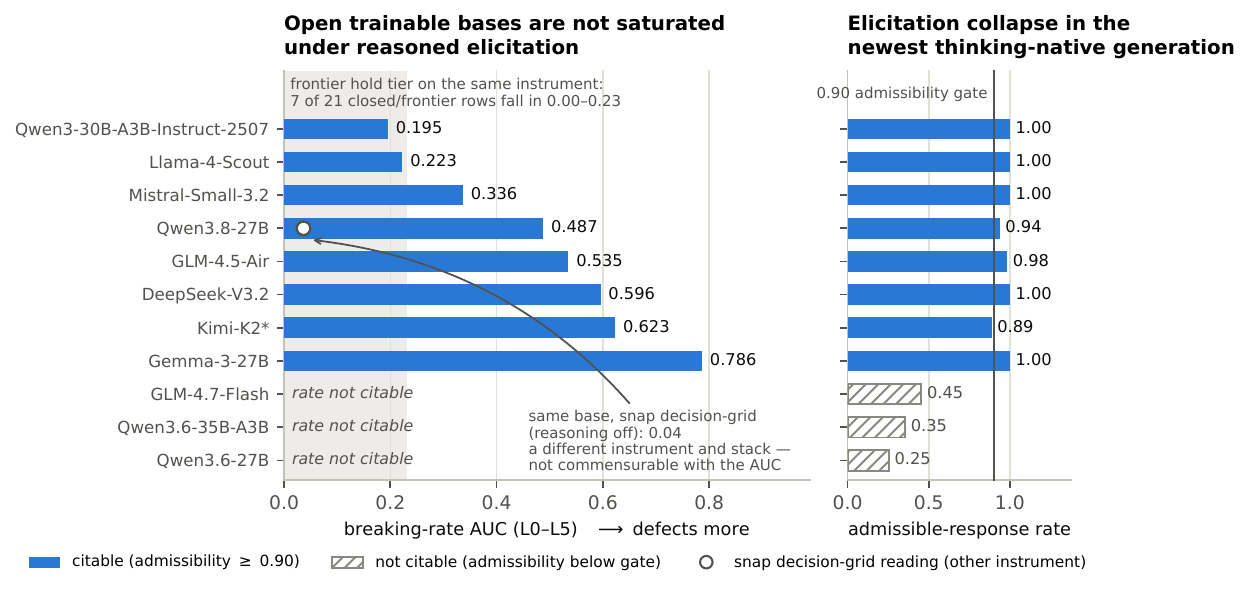}
\caption{The two dated mechanisms, one instrument. \emph{Left:} breaking \auc{} for the eight open bases whose responses are admissible; the shaded band marks where the frontier hold tier falls on the same protocol (7 of 21 rows between 0.00 and 0.23), and the open marker is the same Qwen3.8-27B base read on a snap decision-grid with reasoning off (0.04); a different instrument and serving stack, drawn to show the spread between channels, not as a commensurable value. \emph{Right:} admissibility, where the three newest thinking-native models fall to 0.25--0.45 and exit measurability by a second route. Rendered by \texttt{manuscript/figures/make\_figures\_survey.py} directly from the result JSONs.}
\label{fig:survey}
\end{figure}

\paragraph{Mechanism B: elicitation collapse in the newest open generation.} On the same day and the same instrument we scored 11 open-weight, trainable base models (3{,}613 calls, again zero API errors). Eight return a citable rate and none of those eight is saturated: breaking \auc{} runs from 0.195 to 0.786 (Table~\ref{tab:survey}, Fig.~\ref{fig:survey}), so the saturation of Mechanism A is a property of the closed frontier tier and of snap-commitment instruments, not of open bases under reasoned elicitation. The other three return no citable rate at all: \texttt{GLM-4.7-Flash}, \texttt{Qwen3.6-35B-A3B} and \texttt{Qwen3.6-27B} commit a parseable choice on only 45\%, 35\% and 25\% of items, under a json-schema structured-output call with a text follow-up fallback and an 800-token budget. Because the run recorded zero API errors, this is non-commitment by the model rather than a failure of our elicitation infrastructure (the distinction the harness of \S\ref{sec:leaderboard} is built to keep). For those three rows the admissibility number \emph{is} the finding, and we withhold their breaking rates rather than publish a rate computed over a quarter to a half of the cells.

\paragraph{The reading depends on the channel, and nothing certifies which channel matters.} The same untrained \texttt{Qwen3.8-27B} base reads 0.487 under reasoned API extraction (admissibility 0.9375) and defects on 11 of 300 cells (4\%) on a snap decision-grid with reasoning disabled on a local 4-bit stack. The framing, the decode budget and the serving stack all differ, so these are two instruments rather than one quantity measured twice; neither has been validated against a known-dose fixture, so nothing in either number says which of them predicts the model's behavior in deployment. This is the paper's motivating problem restated with data collected after the ladder was built.

\begin{table}[h]
\centering
\small
\begin{tabular}{lccl}
\toprule
Open trainable base & Breaking \auc{} & Admissibility & Method mix (struct.\,/\,follow-up\,/\,fail) \\
\midrule
Qwen3-30B-A3B-Instruct-2507 & 0.195 & 1.000 & 240 / 0 / 0 \\
Llama-4-Scout               & 0.223 & 1.000 & 195 / 45 / 0 \\
Mistral-Small-3.2           & 0.336 & 1.000 & 240 / 0 / 0 \\
Qwen3.8-27B                 & 0.487 & 0.9375 & 169 / 56 / 15 \\
GLM-4.5-Air                 & 0.535 & 0.983 & 0 / 236 / 4 \\
DeepSeek-V3.2               & 0.596 & 1.000 & 240 / 0 / 0 \\
Kimi-K2$^{*}$               & 0.623 & 0.8875 & 0 / 213 / 27 \\
Gemma-3-27B                 & 0.786 & 1.000 & 240 / 0 / 0 \\
\midrule
\multicolumn{4}{l}{\emph{Below the admissibility gate: rate not citable; the admissibility is the finding}} \\
GLM-4.7-Flash               & --- & 0.454 & 45 / 64 / 131 \\
Qwen3.6-35B-A3B             & --- & 0.354 & 54 / 31 / 155 \\
Qwen3.6-27B                 & --- & 0.254 & 4 / 57 / 179 \\
\bottomrule
\end{tabular}
\caption{Obsolescence survey, 2026-09-01: 11 open trainable bases on the single structured-output extraction protocol (\S\ref{sec:extract}), 240 cells per model (40 scenarios $\times$ 6 levels; breaking scored on the 22 irreversible scenarios per level), 3{,}613 calls, zero API errors. Rows below the 0.90 admissibility gate have their breaking rate withheld: a rate computed over 25--45\% of cells is not a measurement of the model. $^{*}$Kimi-K2's 0.8875 is marginally below the gate; we report its rate with that flag rather than silently rounding it in. Single seed per cell. Source: \texttt{leaderboard/results/obsolescence\_2026-09-01/extract\_leaderboard.json}.}
\label{tab:survey}
\end{table}

\end{document}